\documentclass{article} %
\usepackage{iclr2026_conference,times}

\usepackage{amsmath,amsfonts,bm}

\def\eqref#1{equation~\ref{#1}}

\def\1{\bm{1}}

\DeclareMathAlphabet{\mathsfit}{\encodingdefault}{\sfdefault}{m}{sl}
\SetMathAlphabet{\mathsfit}{bold}{\encodingdefault}{\sfdefault}{bx}{n}

\usepackage{xcolor}
\usepackage{xspace}
\usepackage{amsmath}
\usepackage{graphicx}
\usepackage{booktabs}
\usepackage{array}
\usepackage{makecell}
\usepackage{listings}
\usepackage{algorithm}
\usepackage{algorithmicx}
\usepackage{algpseudocode}
\usepackage{wrapfig}
\usepackage{multirow}
\usepackage{pbox}
\usepackage{caption}
\usepackage{subcaption}
\usepackage{comment}
\usepackage{pdfpages}

\usepackage{inconsolata}   %

\usepackage{tikz}
\usetikzlibrary{positioning, fit, shapes}
\usetikzlibrary{bayesnet}
\usetikzlibrary{arrows}

\newcommand{\ie}{\emph{i.e.,}\ }
\newcommand{\eg}{\emph{e.g.,}\ }

\newcommand{\sect}[1]{\S\ref{#1}}
\newcommand{\sectpar}[1]{(\S\ref{#1})}

\definecolor{lightgray}{RGB}{169, 183, 198} %
\definecolor{darkblue}{RGB}{0, 0, 128} %
\definecolor{darkgray}{RGB}{50, 50, 50} %

\definecolor{lightred}{HTML}{cd5c5c}
\definecolor{green}{HTML}{2ca02c}

\definecolor{gptfour}{HTML}{c94cbe}
\definecolor{random}{HTML}{606264}
\definecolor{ranking}{HTML}{17becf}
\definecolor{gold}{HTML}{1f77b4}
\definecolor{adaptive}{HTML}{9467bd}

\newcommand{\modelname}[1]{\ttfamily{#1}\rmfamily}
\newcommand{\correctanswer}{\ensuremath{a^*}}
\newcommand{\multchoiceq}{\ensuremath{q_{mc}}\xspace}

\newcommand{\unconditionalanswer}{\texttt{Sample\_Answers}\xspace}
\newcommand{\simulate}{\texttt{Simulate\_Student}\xspace}
\newcommand{\infer}{\texttt{Infer\_Misconception}\xspace}
\newcommand{\checkanswer}{\texttt{Check\_Cycle}\xspace}

\newcommand{\mathcommand}[1]{\ensuremath{#1}\xspace}

\newcommand{\uncondanswer}{\mathcommand{a}}
\newcommand{\inferredmisc}{\mathcommand{m}}
\newcommand{\reasonmisc}{\mathcommand{r^{m}}}
\newcommand{\reasonstudenttwo}{\mathcommand{r^{s}}}
\newcommand{\answertwo}{\mathcommand{s}}

\newcommand{\algcomment}[1]{\textcolor{lightgray}{\# {#1}}}

\newcommand{\method}{\textsc{mistake}\xspace}
\newcommand{\methodone}{\textsc{\method-Generate}\xspace}
\newcommand{\methodmath}{\ensuremath{\method\xspace}}
\newcommand{\methodfull}{\method (\textsc{Modeling Incorrect Student Thinking And Key Errors})\xspace}
\newcommand{\methodtwo}{\textsc{\method-Update}\xspace}

\newcommand{\studentmodel}{\ensuremath{M_s}\xspace}
\newcommand{\misconceptionmodel}{\ensuremath{M_m}\xspace}

\newcommand{\methodalg}{\methodtwo}

\newcommand{\starname}{\textsc{STaR}\xspace}

\newcommand{\llamaeightB}{\modelname{Llama-3.1-8B-Instruct}\xspace}
\newcommand{\gpt}{\modelname{GPT}\xspace}
\newcommand{\gptjudge}{\modelname{GPT-4o-mini}\xspace}
\newcommand{\gptthreefive}{\modelname{GPT-3.5-turbo}\xspace}
\newcommand{\gptfourone}{\modelname{GPT-4.1}\xspace}
\newcommand{\gptfouro}{\modelname{GPT-4o}\xspace}

\definecolor{studentcolor}{HTML}{000080}
\definecolor{misccolor}{HTML}{ae12c9}

\newcommand{\studentcolor}{\textcolor{studentcolor}}

\newcommand{\misccolor}{\textcolor{misccolor}}

\newcommand{\textinline}[1]{[\emph{#1}]\xspace}

\newcommand{\methodfullvariant}{\method-\textsc{cycle+correct}\xspace}
\newcommand{\methodcorrectvariant}{\method-\textsc{correct}\xspace}
\newcommand{\methodnocycle}{\textsc{no-cycle}\xspace}
\newcommand{\methodstudentonly}{\textsc{Student-Only}\xspace}
\newcommand{\methodmisconceptiononly}{\textsc{Misconception-Only}\xspace}

\newcommand{\prompt}{\ttfamily\footnotesize}

\usepackage{hyperref}
\usepackage{url}

\title{Learning To Make MISTAKEs: Modeling \\ Incorrect Student Thinking And Key Errors}

\author{Alexis Ross \& Jacob Andreas\\
MIT CSAIL\\
\texttt{\{alexisro,jda\}@mit.edu} \\
}

\iclrfinalcopy %
\begin{document}

\maketitle

\begin{abstract}
    Research on reasoning in language models (LMs) predominantly focuses on improving the correctness of their outputs. But some important applications require modeling reasoning patterns that are \emph{incorrect}. 
    For example, automated systems that can reason about and simulate student errors are useful for providing real-time feedback in the classroom or offline practice for educators-in-training.
    This paper presents a new method, \method, that (1) constructs high-quality synthetic examples of reasoning errors by leveraging cycle consistency between incorrect answers and latent misconceptions; and (2) uses the generated data to learn models for student simulation, misconception classification, and answer generation.
    We evaluate \method on three educational tasks and find that it results in (1) higher accuracy when \emph{simulating incorrect student answers} based on specific misconceptions, (2) increased performance \emph{inferring latent misconceptions} from observed incorrect answers, and (3) higher alignment with expert-written distractor answers when \emph{generating incorrect answers} (\eg for multiple-choice tests). 
    \end{abstract}

\begin{figure}[h!]
    \centering \includegraphics[width=\columnwidth]{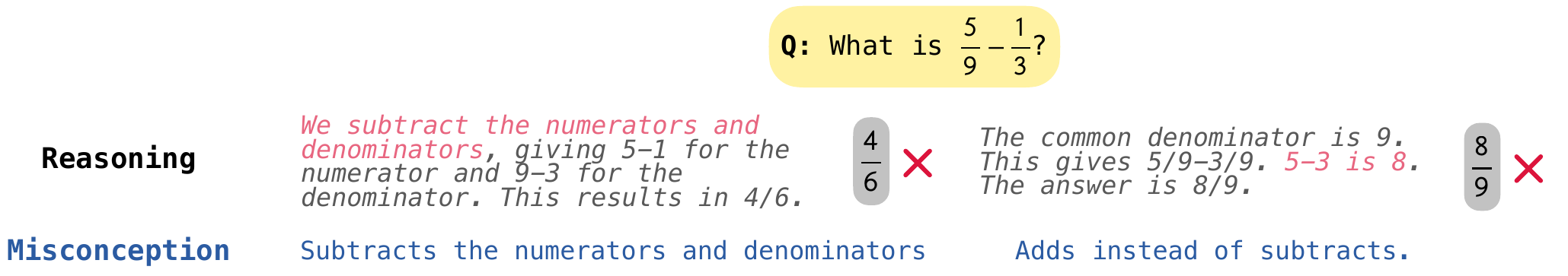} 
    \caption{Examples of mathematical errors that result from common misconceptions shared among students.}
    \label{fig:teaser}
\end{figure}

\section{Introduction}

There is a substantial body of language model (LM) research focused on generating high-quality reasoning traces that lead to correct answers \citep{cot,nye2022show,star}.
However, many applications of LMs require modeling how reasoning can be \emph{wrong}. For example, in education, being able to understand the common reasoning errors that students make allows for tailored assessment and instruction. In addition, recent work has applied LMs to simulate students for uses such as teacher training \citep{gpttraining} and evaluating AI tutors \citep{dict,socraticlm}, both of which require being able to simulate their incorrect reasoning.
Outside of education, work in the social sciences on simulating human behavior with LMs, for example in psychology \citep{dillion2023,demszky2023psychology,park2024generative} and economics \citep{horton2024}, also requires being able to model cognitive biases and fallacies.

Figure~\ref{fig:teaser} shows examples of human-like incorrect reasoning in an elementary mathematics setting. The figure gives examples of two errors in solving a question about fractions; these particular errors result from specific misconceptions shared by many learners encountering fraction arithmetic for the first time. 
Modeling such errors requires a nuanced understanding of the relationship between mathematical concepts and how people reason about them. As we show, current LMs are much worse at simulating such errors than they are at performing correct reasoning to, \eg solve math problems.

In this paper, we introduce an \emph{unsupervised} procedure for generating high-quality, human-like reasoning data similar to what is shown in Figure~\ref{fig:teaser}. The key idea behind our approach is to leverage cycle consistency between incorrect answers and their underlying misconceptions; this allows us to augment a set of questions with misconceptions, reasoning, and incorrect answers without requiring {any} examples of human-generated errors. We then use this data to improve performance on three education tasks. We refer to the end-to-end method as \textbf{\methodfull}. 

\method is built from two procedures. The \textbf{inner loop}, \methodone, samples plausible triples (misconception, faulty reasoning, answer) by decoding from a model with a cycle consistency constraint. The \textbf{outer loop}, \methodtwo, fine-tunes models on the cycle consistent data. Together, they provide an end-to-end, unsupervised procedure for generating large numbers of synthetic reasoning traces with interpretable errors; they additionally yield both a \textbf{student simulation} model capable of simulating \emph{reasoning with misconceptions}, and a \textbf{misconception inference} model that can observe a student's behavior and \emph{reason about misconceptions} to identify what the student is confused about.

Models trained via \method achieve {improved performance} on three education tasks:

\begin{enumerate}
    \item \textbf{Student Simulation:} Given a misconception, this task requires simulating the incorrect reasoning and answer that a student would produce. \method improves accuracy by up to \textbf{9\%} \sectpar{sec:results_student_simulation}. 
    \item \textbf{Misconception Inference:} This task involves inferring a student's misconception based on an incorrect answer they provided. \method leads to a~\textbf{15\%} improvement in performance on this task \sectpar{sec:results_misconception_inference}. 
    \item \textbf{Distractor Generation:} 
    This task evaluates \method's ability to generate high-quality {incorrect distractor answers}. 
    \method generates distractor answers that are more often found in the expert-written distractor choices for each question, with a \textbf{64.6\%} increase in precision, suggesting that \method generates incorrect data that is more human-like \sectpar{sec:results_distractor_generation}.
\end{enumerate}

Together, our results highlight the promise of explicitly modeling patterns of incorrect reasoning across a range of educational domains.\footnote{Our code is publicly available at \url{https://github.com/alexisjihyeross/mistake}.}

\begin{figure*}[ht!]
    \centering \includegraphics[width=\textwidth]{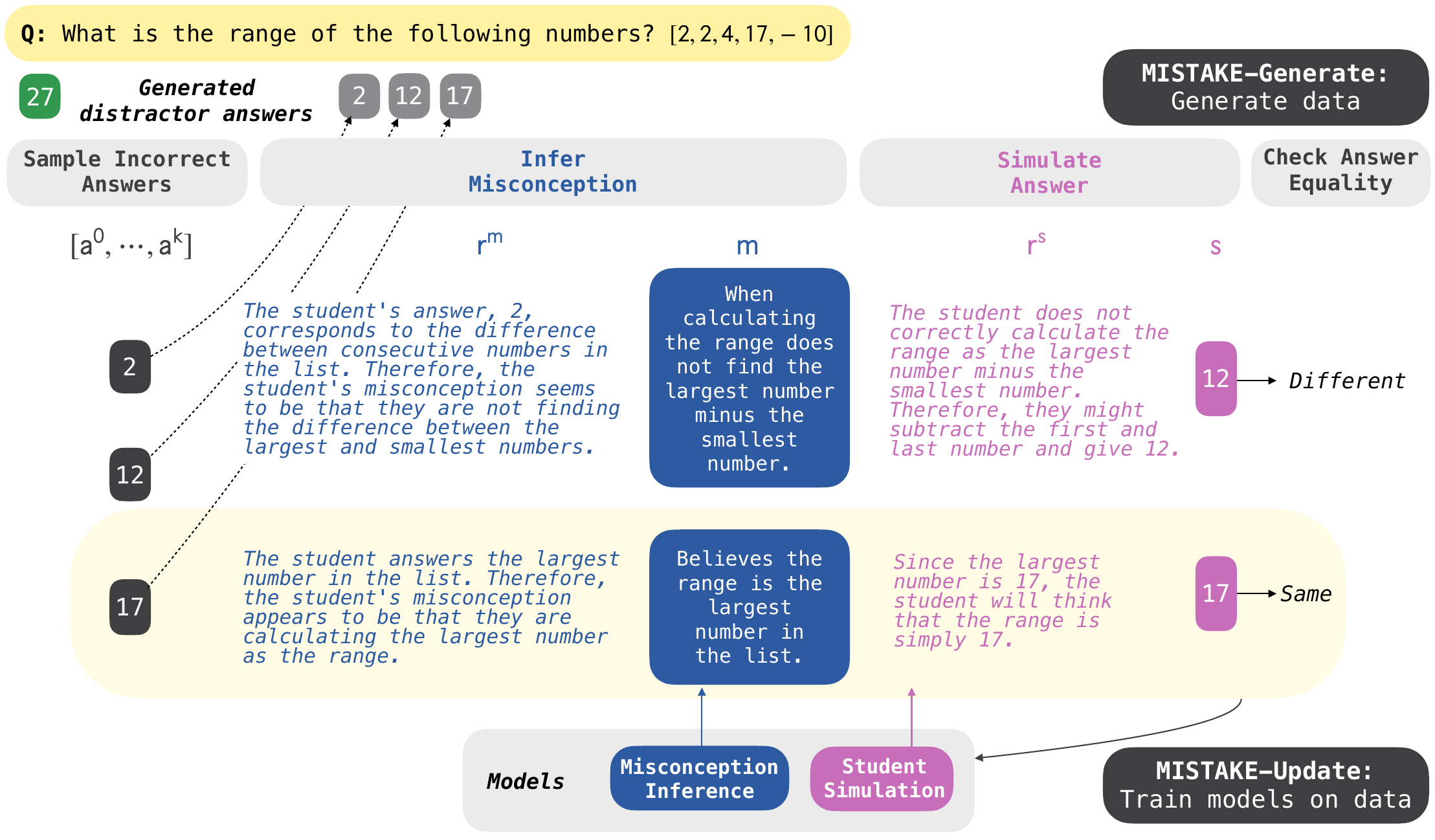} 
    \caption{Overview of \method. \methodone generates data by enforcing cycle consistency between misconceptions, reasoning traces, and answers. \methodtwo iteratively trains student simulation and misconception inference models on this data, generates new data using \methodone and these models, and repeats.
    }
    \label{fig:method}
\end{figure*}

\section{Related Work}
\label{s:related}

\paragraph{Education}
Work on modeling student misconceptions has a long history in education research \citep{brown1978diagnostic,vanlehn1990mindbugs,feldman2018}, and more recently within AI for education. In a synthetic evaluation framework, \citet{ross-andreas-2024-toward} find that LLMs can infer student misconceptions and adapt teaching strategies better than simple baselines but worse than more sophisticated methods that explicitly model misconceptions. Similarly, \citet{scarlatos2024knowledge} find that combining LMs with knowledge tracing (KT) leads to better estimates of student knowledge states than KT-only methods in dialogue settings. \citet{malalgoqa} find that LLMs are much worse at identifying incorrect reasoning containing misconceptions than they are at identifying correct reasoning. All of these studies suggest that there is headway to be made in using LMs to explicitly model student misconceptions.

A key challenge in such research is the scarcity of high-quality data, particularly expert-annotated examples of real student misconceptions. {The DrawEduMath dataset contains students' hand-written solutions annotated by expert teachers \citep{DrawEduMath}; however, while it contains annotations of students' errors and strategies used to solve the problem, it lacks standardized annotations of higher-level misconceptions}; similarly, the MalAlgoQA dataset contains math problems with associated incorrect answers and incorrect rationales, but the incorrect rationales are again problem-specific \citep{malalgoqa}.\footnote{For example, an incorrect rationale in the MalAlgoQA dataset is: ``Chose the number of times a star is picked in the 1st 50 cards drawn.'' This is an incorrect reasoning step specific to a particular problem, not reflective of the kinds of higher-level misconceptions that affect student reasoning across math problems.} The EEDI Mining Misconceptions in Mathematics dataset \citep{eedi} is one of a few datasets that contain natural student data with annotations of generalizable error descriptions. 
However, the process of collecting expert teacher annotations remains resource-intensive, limiting the scalability of these datasets.

In light of these data limitations,
 recent works have used off-the-shelf LMs to simulate students.
 Recent tutoring benchmarks use LM-simulated students for both dataset construction and evaluation  \citep{mathdial,stepverify,socraticlm,dict}. Existing approaches predominantly aim to simulate general student performance or skills rather than specific misconceptions \citep{lu2024, benedetto2024}.
While \citet{sonkar2024llmbasedcognitivemodelsstudents} propose a Python library that models misconceptions in linear algebra, their approach, based on a hand-engineered graphical model, is limited to specific types of equations.
In contrast to this past work, \method provides an unsupervised method for generating high-quality data with misconceptions and learning models from this data that can simulate misconceptions in a natural educational domain. %

Outside of student simulation, another promising educational application of AI is in helping automate \emph{assessment}, \eg by 
constructing high-quality distractor answers for multiple-choice questions. Previous work has leveraged in-context learning with nearest-neighbor examples \citep{mcnichols2024automateddistractorfeedbackgeneration,feng2024distractor}. \citet{scarlatos2024distractor} introduce a ranking model to predict student selection probabilities for distractors, using this to filter LM generated options, and \citet{divert} introduce a method that jointly learns textual descriptions of the errors behind incorrect answers along with the incorrect answers. However, all of these methods require a dataset of existing distractors to use as candidates/training examples. As we will see, \method produces high-quality distractors as a byproduct of training, \emph{without} a dataset of existing human-authored distractors.

\paragraph{Reasoning}  

Our work is also related to the literature on learning to reason
\citep{cot,nye2022show,li-etal-2023-language-modeling,star,zelikman2024quietstar,deepseek2025}. Most closely related is \starname, an algorithm that iteratively samples reasoning traces from a model, trains on a filtered set of traces, re-samples, and repeats \citep{star}. 
Unlike \starname, which assumes access to target labels, \method is unsupervised and learns to impute both reasoning and target (incorrect) labels.

\section{\methodfull}
\label{sec:method}
Our ultimate goal is to train two distinct models: first a \textbf{student simulation model}  $\studentmodel$ that can generate plausible student behavior \emph{conditioned} on student descriptions (which may include misconceptions); second a \textbf{misconception inference model} $M_m$ that can observe a student trace and likely sources of student errors.
\method trains these models via two nested procedures: an inner loop \methodone (\sect{sec:method_data_construction}) that generates data by enforcing cycle consistency between inferred misconceptions, generated reasoning traces, and answers; and an outer loop \methodtwo (\sect{sec:method_alg}) that uses the data to finetune 
$\studentmodel$ and $M_m$.
Figure~\ref{fig:method} shows an overview of \method  with examples.

\subsection{\methodone: Unsupervised Data Construction}
\label{sec:method_data_construction}

Algorithm~\ref{alg:datagen} presents an overview of \methodone, which uses an existing base LM $M$, student model $\studentmodel$, and misconception model $M_m$ to generate new traces exhibiting reasoning with misconceptions.
Below we explain how the procedure works step-by-step.
\begin{algorithm*}[h]
\caption{\methodone: Unsupervised Data Generation}
\textbf{Input:} Questions $Q$, pretrained model $M$, \studentcolor{student simulation model $\studentmodel$}, \misccolor{misconception inference model $\misconceptionmodel$}
\begin{algorithmic}[1]
\For{each question and correct answer pair $(q, \correctanswer) \in Q$}
    \State $[\uncondanswer_0, \uncondanswer_1, \uncondanswer_2] \leftarrow \unconditionalanswer(q, \correctanswer, M)$ \algcomment{Sample 3 incorrect answers with $M$}
    \State $\multchoiceq \leftarrow (q, \uncondanswer_0, \uncondanswer_1, \uncondanswer_2, \correctanswer)$ \algcomment{Create a multiple choice question}
    \For{each incorrect answer $\uncondanswer$}
        \State $\reasonmisc, \inferredmisc \leftarrow \misccolor{\infer(\multchoiceq, \uncondanswer, \misconceptionmodel)}$ \algcomment{Infer misconception with $\misconceptionmodel$}
        \State $\reasonstudenttwo, \answertwo \leftarrow \studentcolor{\simulate(\multchoiceq, \inferredmisc, \studentmodel)}$ \algcomment{Simulate student based on \inferredmisc with $\studentmodel$}
        \State $w \leftarrow \begin{cases} \alpha & \text{if }\checkanswer(\uncondanswer, \answertwo, \correctanswer,M) \\ 1 & \text{otherwise} \end{cases}$ \algcomment{Check cycle consistency with $M$}
        \State Add $(\multchoiceq, \reasonstudenttwo, \answertwo, \reasonmisc, \inferredmisc, w)$ to dataset $D$
    \EndFor
\EndFor
\State \Return Dataset $D$ of weighted examples
\end{algorithmic}
\label{alg:datagen}
\end{algorithm*}

\paragraph{\unconditionalanswer} The first step in \methodone is to sample a set of incorrect answers $[\uncondanswer_0, \cdots, \uncondanswer_k]$ that a student might have when solving a question $q$. We sample these answers by prompting a pretrained LM $M$, conditioning on the question $q$ and the correct answer $\correctanswer$. The generated answers are used as (a) distractors for the student simulation module \simulate, which takes in multiple-choice questions, and (b) as candidate labels for the misconception inference module \infer module and rest of the \methodone process. For example, for the question shown in Figure~\ref{fig:method}, \textinline{What is the range of the following numbers? [2, 2, 4, 17, -10]}, \unconditionalanswer may output $[2, 12, 17]$.

\paragraph{\misccolor{\infer}} Given the multiple choice question $\multchoiceq$ with generated distractor answers and specific candidate answer $\uncondanswer$, the \infer module uses the misconception model $M_m$ to infer the conceptual misunderstanding that would have led to the incorrect answer $\uncondanswer$. The outputs of \infer are the inferred misconception \inferredmisc, along with a reasoning trace \reasonmisc explaining how it arrived at that conclusion. For example, for candidate answer $\uncondanswer = 17$, \infer may output \reasonmisc = \textinline{The student answers the largest number in the list. Therefore, the student's misconception appears to be that they are calculating the largest number as the range} and \inferredmisc = \textinline{Believes the range is the largest number in the list}.

\paragraph{\studentcolor{\simulate}} Given a question \multchoiceq and inferred misconception \inferredmisc, \simulate uses the student simulator \studentmodel to simulate the step-by-step reasoning and final answer that a student would produce if they had the misconception. 
For example, for misconception \inferredmisc = \textinline{Believes the range is the largest number in the list}, \simulate may output \reasonstudenttwo = \textinline{The student answers the largest number in the list. Therefore, the student's misconception appears to be that they are calculating the largest number as the range} and \answertwo = 17.

\paragraph{\checkanswer} The cycle consistency check
serves as an unsupervised quality filter. If \checkanswer returns true, this provides strong evidence that the inferred misconception \inferredmisc has the desired relationship with the original answer \uncondanswer. This is because if the misconception were incorrect or unrelated to the answer it would be unlikely that simulating a student with that misconception would produce the same answer again. For example, the first misconception in Figure~\ref{fig:method}, \textinline{Believes the range is the largest number in the list}, is a high-quality misconception and, when simulated faithfully, should lead to the original answer $a=17$. The cycle consistency check therefore verifies both directions of the relationship: that the misconception explains the original answer (answer $\rightarrow$ misconception) and that the misconception leads back to the same answer (misconception $\rightarrow$ answer). Examples that pass this check are given higher weight ($w = \alpha$) in the training data, as they represent more reliable examples of the relationship between misconceptions and incorrect answers.

There are some boundary cases for the cycle consistency check. For example, the second misconception \textinline{When calculating the range does not find the largest number minus the smallest number} is too general to be able to re-simulate the exact original sampled answer $s=2$, as it could explain many incorrect answers. However, we may still want to include the re-simulation \textinline{The student does not... Therefore, they might subtract the first and last number and give 12} since it may still be useful for learning how to generally simulate student mistakes, as long as it leads to an incorrect answer. 
For this reason, we
explore two variants of \method \sectpar{ssec:method_variants}: one that filters misconceptions based on the \emph{strong} constraint that the inferred misconception results in the same incorrect answer that was sampled (\ie $\answertwo = a$), which we call \methodfullvariant, and another that uses the \emph{weaker} constraint that the simulated answer is not the correct answer (\ie $\answertwo \neq \correctanswer$), which we call \method-\textsc{cycle}.

\subsection{\methodtwo: Iterative Training Algorithm}
\label{sec:method_alg}

\methodtwo is an iterative algorithm that trains two models on related tasks using the data generated by \methodone as described in \sect{sec:method_data_construction}. Algorithm~\ref{alg:method} summarizes the iterative training process used to train the \studentcolor{student simulation} model $\studentmodel$ and the \misccolor{misconception inference} model $\misconceptionmodel$.

We subset the data generated by \method into two datasets: one for training a student simulation model $\studentmodel$ and one for training a misconception inference model $\misconceptionmodel$. $\studentmodel$ is trained on the simulated incorrect answers \answertwo and reasoning traces \reasonstudenttwo used to generate those answers, while $\misconceptionmodel$ is trained on the incorrect answers \answertwo and inferred misconceptions \inferredmisc. 

\begin{algorithm*}[h]
\caption{\methodtwo: Iterative Training of Student Simulation and Misconception Inference Models}
\textbf{Input:} a pretrained language model $M$
\begin{algorithmic}[1]
\State $D_0 \leftarrow \methodmath(M, M)$ \algcomment{Generate initial dataset with \method using $M$}
\State \studentcolor{$D_0^s \leftarrow \{(x=(q, \inferredmisc), y=(\reasonstudenttwo, \answertwo)) \mid (q, \reasonstudenttwo, \answertwo, \reasonmisc, \inferredmisc) \in D_0\}$} \algcomment{Student simulation data}
\State \misccolor{$D_0^m \leftarrow \{(x=(q, \answertwo), y=(\reasonmisc, \inferredmisc)) \mid (q, \reasonstudenttwo, \answertwo, \reasonmisc, \inferredmisc) \in D_0\}$} \algcomment{Misc. inference data}

\For{$t = 1$ to $T$}
    \State \studentcolor{$\studentmodel \leftarrow \text{train}(M, D_{t-1}^s)$} \algcomment{Finetune orig model on new student simulation data}
    \State \misccolor{$\misconceptionmodel \leftarrow \text{train}(M, D_{t-1}^m)$} \algcomment{Finetune orig model on new misconception inference data}
    \State $D_t \leftarrow \methodmath(\studentmodel, \misconceptionmodel)$ \algcomment{Generate new \method data with finetuned $\studentmodel$, $\misconceptionmodel$}
    \State \studentcolor{$D_t^s \leftarrow \{(x=(q, \inferredmisc), y=(\reasonstudenttwo, \answertwo)) \mid (q, \reasonstudenttwo, \answertwo, \reasonmisc, \inferredmisc) \in D_t\}$} \algcomment{Student simulation data}
    \State \misccolor{$D_t^m \leftarrow \{(x=(q, \answertwo), y=(\reasonmisc, \inferredmisc)) \mid (q, \reasonstudenttwo, \answertwo, \reasonmisc, \inferredmisc) \in D_t\}$} \algcomment{Misc. inference data}
    \EndFor
\State \Return \studentcolor{$\studentmodel$}, \misccolor{$\misconceptionmodel$} \algcomment{Return trained models}
\end{algorithmic}
\label{alg:method}
\end{algorithm*}

Inspired by \starname \citep{star} and other expectation-maximization-style algorithms for training LMs \citep[\eg][]{bostrom2024language}, we iteratively finetune $\studentmodel$ and $\misconceptionmodel$ on the data generated by \methodone, using the finetuned models to generate new data, and repeating.
\methodtwo seeds the iterative process by using a pretrained LM $M$ as $\studentmodel$ and $\misconceptionmodel$ to generate the initial dataset $D_0$. After the first iteration, the finetuned models are used to generate the next round of data with \methodone, which is used to finetune the models again. This process repeats for $T$ iterations. {The final results are trained $M_s$ and $M_m$ models useful for simulating student reasoning and inferring misconceptions respectively. Importantly, both $M_s$ and $M_m$ are reasoning models---in contrast to existing EM-style training procedures for LMs, both the inference model and the forward simulation model ``think out loud'' and improve their behavior over time.}

\section{Experiments}
\label{sec:experiments}
In this section, we describe our experiments evaluating \method on three education tasks.

\subsection{Data}
We work with the EEDI Mining Misconceptions in Mathematics dataset, which consists of 1,857 K--12 math questions \citep{eedi}. Each question has four expert-written multiple choice options that correspond to misconceptions that a student might have.\footnote{Of the 7,428 total answer choices in the dataset, 4,338 of them are labeled with text descriptions of corresponding misconceptions. There are 2,587 unique misconceptions in the dataset.}
The incorrect answer choices and misconception annotations in EEDI are written by expert educators. We evaluate on these labels to determine whether \method, which only ever trains models on synthetically generated misconception data, generalizes to \emph{real-world} data.

We subset the EEDI data into train (70\%), validation (15\%), and test splits (15\%) by holding out math {questions} so that all (question, misconception, answer) pairs for the same question end up in the same split. We report results on the test set unless otherwise specified. 

\subsection{Tasks}
\label{ssec:tasks}

We evaluate \method on three tasks that are useful for tailoring assessment and instruction to different students and providing offline practice for educators-in-training.

\paragraph{Student Simulation} We evaluate a model's ability to simulate the incorrect answer that a student with a particular misconception would give. For each incorrect multiple choice answer in EEDI that has a labeled misconception, we evaluate whether the incorrect answer generated by the student simulation model, conditioned on a misconception description, is the same as the ground truth incorrect answer corresponding to the misconception. We evaluate the \textbf{accuracy} of simulated answers through pattern matching on generated letters corresponding to answer choices. 

\paragraph{Misconception Inference} We also run the evaluation in the reverse direction: We evaluate the misconception inference model's accuracy at predicting a student's latent misconception from the incorrect answer they gave. Given a math question, an incorrect multiple choice answer, and a ground-truth misconception associated with the incorrect answer, we prompt the misconception inference model to output a description of the misconception that would lead to the answer. To evaluate the generated misconception, we embed the generated misconception, ground truth misconception, and full list of possible misconceptions in the EEDI data.  We use the \modelname{Instructor-XL} model to embed misconceptions \citep{instructor}.\footnote{The instruction for the \modelname{Instructor-XL} embedding model is: \textinline{Represent the following misconception that a student might have in solving K-12 math problems for retrieving similar misconceptions.}} We then sort the list of candidate misconceptions by their cosine similarity to the generated misconception and evaluate the mean average precision at k, or \textbf{MAP@k} score, a metric introduced in the challenge along with the EEDI data:
\[ \text{MAP@k} = \begin{cases} 
\frac{1}{p} & \text{if true misconception found at}\\ & \text{position p in top k misconceptions} \\
0 & \text{otherwise}
\end{cases} \]
where \( p \) is the position where we find the true misconception in our sorted list of predictions. 
For example, if the true misconception appears at position 3 in our sorted list, then the score would be $\frac{1}{3}$. If the true misconception is not found in the top k predictions, the score is 0. We report results for k=25, as this is the value used by the EEDI Mining Misconceptions in Mathematics Challenge.\footnote{The challenge can be found at: \url{https://www.kaggle.com/competitions/eedi-mining-misconceptions-in-mathematics}}

\paragraph{Distractor Generation} We evaluate the ability of \method to generate human-aligned distractor answers. We measure the \textbf{precision} of generated distractor answers that match expert-written incorrect answers after filtering for cycle-consistency. For each (generated distractor, ground-truth distractor answer) pair, we prompt a judge LM (\gptjudge) to determine whether they are equal (see Table~\ref{tab:gpt_student_judge} for the prompt). In a manual analysis of the \gptjudge judge's annotations, we found that they were 100\% accurate.\footnote{We validate the accuracy of the \gptjudge judge by manually annotating 40 randomly sampled judgments of whether a generated distracted answer choice is the same as a ground truth answer choice. We find that all 40 answer judgments are correct. This high accuracy is explained by this judgment task being easy: The model simply needs to judge whether two answers are the same answer in different forms (\eg recognizing that the answer “Neither Tom nor Katie are correct” is the same as the answer “Neither is correct”), and therefore the GPT-4o-mini model can suffice for this task.}
We then compute the proportion of distractor answers that are judged to be the same as at least one of the ground truth incorrect answers for the question.

\subsection{Method Variants} 
\label{ssec:method_variants}

We experiment with several variants of \method that differ in \checkanswer conditions. Table~\ref{tab:variants} summarizes the different variants.
The first is \textbf{\methodfullvariant}, which uses the full cycle consistency criterion. In particular, \methodfullvariant \emph{upweights} examples where the generated answer is fully cycle consistent---\ie the same as the answer sampled with \unconditionalanswer (\ie $s=a$)---and \emph{removes} examples where the generated answer equals the correct answer, \ie $s = \correctanswer$.\footnote{We experimented with removing all examples that were not cycle consistent rather than upweighting ones that were, but found that this led to slightly worse results.}
The second variant is \textbf{\methodcorrectvariant}, which only removes examples where the generated answer equals the correct answer, \ie $s = \correctanswer$.
The last variant is \textbf{\methodnocycle}, which ablates both types of cycle consistency conditions and weights all examples equally.

We also ablate the joint training of student simulation and misconception inference models by only training one of the two models, holding the other fixed. We refer to these ablations as \textbf{\methodstudentonly} and \textbf{\methodmisconceptiononly}.

\subsection{Experimental Set-Up}
\label{s:experimental_details}
For all \method variants and ablations, \llamaeightB \citep{llama} is our base model. This model serves as the backbone for all five steps in \method and in \methodalg. {We prompt all models with few-shot examples with manually written reasoning traces.} See the Appendix for details. We run 5 random seeds per experiment.

In addition to the unsupervised quality filters described in \sect{ssec:method_variants}, we filter examples where the generated data consists of empty strings, which happens if the model does
not generate an output in the correct format.\footnote{We remove examples where \reasonstudenttwo or \answertwo are empty strings from $D^s$, and we remove examples where \reasonmisc or \inferredmisc are empty strings from $D^m$.}

For \methodalg, we fine-tune models using LoRA \citep{lora} with rank $r=8$ for up to 4 epochs, with early stopping based on validation loss on the synthetically generated validation dataset. We run experiments for $T=4$ iterations.\footnote{We train all models on a single H100 GPU.}

\begin{figure}[t!]
    \centering \includegraphics[width=\columnwidth]{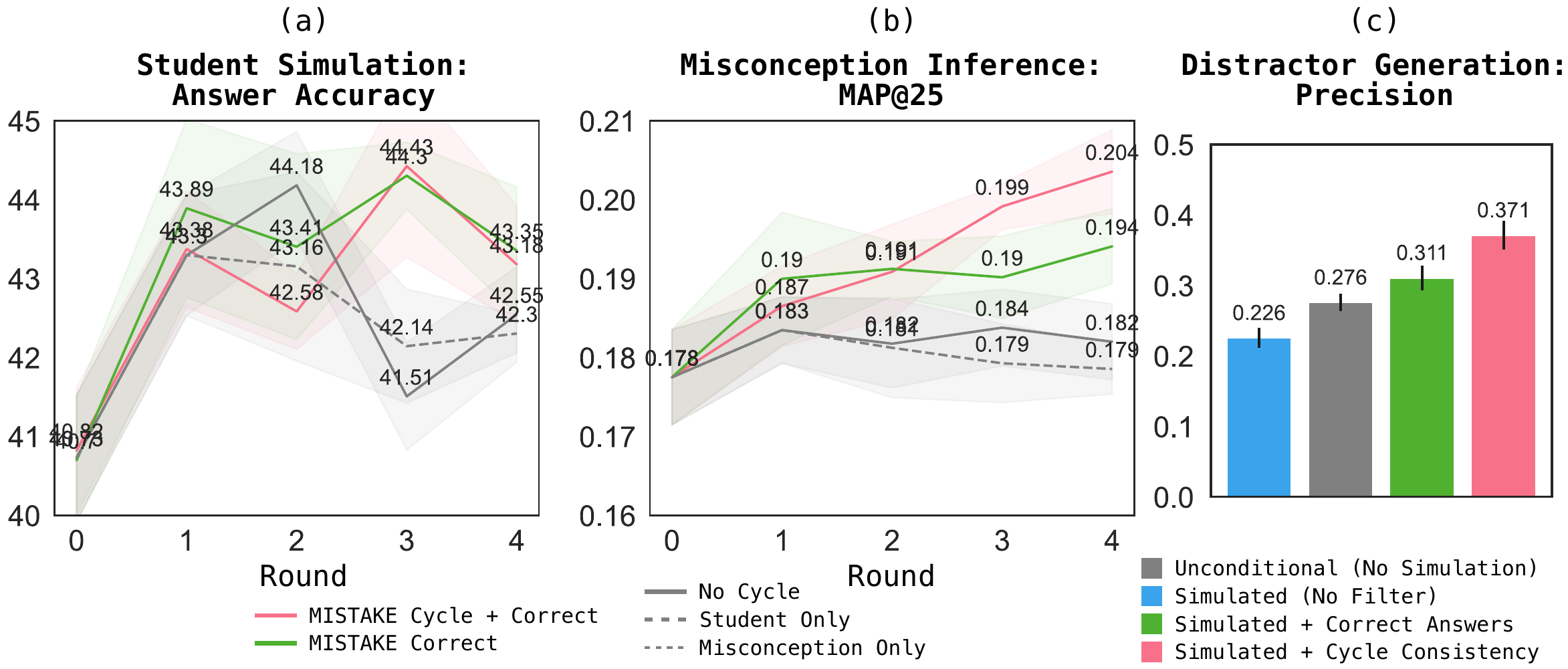} 
    \caption{Results on the three educational tasks described in \sect{ssec:tasks}. We report means and standard errors across 5 random seeds. (a) Student simulation accuracies of \method variants \sectpar{sec:results_student_simulation} (test set). (b) Misconception inference results for \method variants (test set) \sectpar{sec:results_misconception_inference}. (c) Precision of generated distractor answers for \methodfullvariant (validation set) \sectpar{sec:results_distractor_generation}. 
    }
    \label{fig:performance}
\end{figure}

\section{Results}

Figure~\ref{fig:performance} shows how \method variants and ablations perform across training rounds. We provide more detailed presentations of results for each task in the rest of the section. Tables~\ref{tab:examples_student_sim} and \ref{tab:examples_misconception_inference} contain examples of model outputs for the student simulation and misconception inference tasks, respectively.

\subsection{API model references}

Tables~\ref{tab:prompt_comparison} and ~\ref{tab:prompt_comparison_distractors} show how the best results achieved by a \method variant compares to prompting closed \gpt models. We note that these prompted methods are not baselines in that \method could be applied on top of any existing model (as long as it is open); however, they are useful reference points for how frontier LMs perform on these tasks.
Overall, we find that for student simulation and misconception inference, the best performing \llamaeightB models trained with \method perform comparably or better than \gptthreefive for student simulation and misconception inference, and approach the performance of models several orders of magnitude larger. 

Because the cycle-consistency filtering procedure in \method-\textsc{generate} can be applied before fine-tuning, we can also apply it directly to API models. Here we find that \textsc{mistake} improves the precision of generated distractor across scales, including \gptfouro and \gptfourone models.

\begin{table*}[b]
\centering
\footnotesize
\begin{tabular}{lccc}
\hline
\textbf{Model} & 
\shortstack{\textbf{Task} \\{Accuracy (\%)} \\ {}} & 
\shortstack{\textbf{Student} \\ \textbf{Simulation} \\ {Accuracy (\%)}} & 
\shortstack{\textbf{Misconception} \\ \textbf{Inference} \\ {MAP@25}} \\
\hline
\method & $69.4^{\dagger}$ & $44.4$ & $0.204$ \\
\gptthreefive & $74.1$~ & $40.6$~ & $0.206$~ \\
\gptfouro   & $85.0$~ & $64.1$~ & $0.259$~ \\
\gptfourone  & $92.4$~ & $66.3$~ & $0.271$~ \\
\hline
\end{tabular}
\caption{Comparison of the best models trained with \method with results from larger, closed-source \gpt models. $^{\dagger}$Indicates that the result is reported from the pretrained \llamaeightB model. All other results are the best values achieved by a \method variant on the test set (see Figure~\ref{fig:performance} for full performance across rounds).
}
\label{tab:prompt_comparison}
\end{table*}

\begin{table*}[h]
\centering
\footnotesize
\begin{tabular}{lccc}
\hline
\textbf{Model} & 
\textbf{Base model} & 
\textbf{+\method-\textsc{generate}} \\
\hline
\llamaeightB & $0.226$ 
& $\mathbf{0.371}$ ~ $(+0.145)$ \\
\gptthreefive & $0.320$ 
& $\mathbf{0.375}$ ~ $(+0.055)$ \\
\gptfouro   & $0.427$ 
& $\mathbf{0.497}$ ~ $(+0.070)$ \\
\gptfourone  & $0.447$ 
& $\mathbf{0.490}$ ~ $(+0.043)$\\
\hline
\end{tabular}
\caption{Comparison of distractor precision for a variety of base models, with and without the cycle consistency filter condition in \method. See Figure~\ref{fig:performance} for results for other filtering conditions.}
\label{tab:prompt_comparison_distractors}
\end{table*}

\subsection{Student Simulation} 
\label{sec:results_student_simulation}

We find that all models achieve much lower accuracy on student simulation than for the task itself (solving the math questions correctly); the drop in accuracy ranges from \textbf{24.6\%} ({92.4\%} $\rightarrow$ {66.3\%}) to \textbf{45.2\%} (74.1\% $\rightarrow$ 40.6\%). Even powerful LMs such as \gptfouro and \gptfourone struggle to simulate incorrect student answers.
The pretrained \llamaeightB model performs poorly on the student simulation task, with a starting accuracy of \textbf{40.83\%}, which is \textbf{58.8\%} of the model's performance at the task of solving math problems. This difference suggests that student simulation is a more difficult task for current models than solving math correctly.

As shown in Figure~\ref{fig:performance}a, we find that all \method variants lead to some accuracy improvements, but the methods with some version of cycle consistency---\methodfullvariant and \methodcorrectvariant---improve the most. The worst-performing variants are \methodnocycle and \methodstudentonly. The best variant, \methodfullvariant, improves by $\sim$\textbf{9\%} (\textbf{40.83\%} $\rightarrow$ \textbf{44.43\%}).

\subsection{Misconception Inference}
\label{sec:results_misconception_inference}
We see similar trends for the misconception inference task as we do for student simulation. As shown in Figure~\ref{fig:performance}b, we find all \method variants lead to improvements in the MAP@k score, with \methodfullvariant leading to the best performance (\textbf{0.178} $\rightarrow$ \textbf{0.204}, representing a $\sim$\textbf{15\%} improvement over the pretrained model. Again, we find that only training the misconception model, \ie \methodmisconceptiononly, leads to the worst performance.

\subsection{Distractor Generation} 
\label{sec:results_distractor_generation}

Figure~\ref{fig:performance}c shows the precision of generated distractor answers for each question in the validation dataset for models trained on the \methodfullvariant data. We compare multiple sets of generated distractor answers. \textbf{\textsc{Unconditional}} evaluates the answers generated by \unconditionalanswer in \methodone.
We also evaluate the answers output by \simulate in \methodone: \textbf{\textsc{Simulated (No Filter)}} evaluates all of the generated answers. \textbf{\textsc{Simulated + Correct Answers}} only evaluates answers that are not equal to the correct answer, while \textbf{\textsc{Simulated + Cycle Consistency}} is the full cycle consistency condition in \methodone, \ie only evaluating answers that are the same as original sampled answers.

We find that the {simulated} methods with filtering outperform \textsc{Unconditional} and \textsc{Simulated (No Filter)} methods, suggesting that the procedure in \methodone of inferring misconceptions and simulating answers is effective at generating high-quality distractor answers. The distractors generated by \textsc{Simulated + Cycle Consistency} are consistently the most aligned with the ground truth distractors than the other methods, suggesting that the cycle consistency check in particular is an effective way of improving the quality of generated distractors. 
The biggest improvement in distractor precision, with \textsc{Simulated + Cycle Consistency} leading to a \textbf{64.6\%} improvement over \textsc{Unconditional} (\textbf{22.56\%} $\rightarrow$ \textbf{37.14\%}).

In addition, as shown in Table~\ref{tab:prompt_comparison_distractors}, applying the full \textsc{Simulated + Cycle Consistency} filter in \methodone leads to improvements in distractor precision across all models we evaluate, including the most powerful models \gptfouro and \gptfourone.

\section{Conclusion}
\label{sec:conclusion}

Overall, our experiments demonstrate that \method is an effective approach for modeling incorrect reasoning and that it leads to improved performance on three educational tasks, student simulation \sectpar{sec:results_student_simulation}, misconception inference \sectpar{sec:results_misconception_inference}, and distractor generation \sectpar{sec:results_distractor_generation}. We show that the cycle consistency check in \methodone and the joint training of student simulation and misconception inference models in \methodtwo are both key components of this procedure. 
Taken together, these results highlight that while modeling incorrect reasoning is challenging for existing models, \method is an effective first step towards this goal. Future work can explore how the models trained by \method can be used downstream in educational applications, \eg in conjunction with chat-based LLMs to provide tutoring tailored to misconceptions. Another interesting direction for future work is to explore how the cycle consistency conditions in \method can be used to create better user simulators in other settings such as chat-based tutoring or even non-educational domains where users' behaviors may be explained by misconceptions or latent cognitive patterns.

\subsubsection*{Acknowledgments}
This work was supported by the National Science Foundation under grant IIS-2212310 and by the MIT Quest for Intelligence. AR is additionally supported by a National Science Foundation GRFP 2023357727, and JA by a Sloan Research Fellowship. We thank Linlu Qiu and Ben Lipkin for conversations early in the project, as well as Gabriel Grand, Leshem Choshen, and members of the Language and Intelligence group at MIT for feedback on the draft.

\bibliography{custom}
\bibliographystyle{iclr2026_conference}

\newpage
\appendix

\begin{table*}[t!]
    \renewcommand{\arraystretch}{0.8}
    \begin{tabular}{p{0.95\textwidth}}
        \toprule
        \prompt You are an AI assistant designed to judge whether two answers to a math problem are conceptually the same or different. You will be shown a K-12 math problem and two answers to the problem. Please output ‘Same’ if the two answers are conceptually equivalent and just formatted differently. Please output ‘Different’ if the two answers are conceptually different. If one of the answers is the empty string and the other is not, you should output ‘Different’. Please do not actually solve the problem. You should only judge whether the two answers are the same or different. Your output should be formatted as follows:\\\\
        \prompt Judgment: [Same/Different]\\\\
        \prompt Here are several examples to guide your answer:\\\\
        \prompt Math Problem:```Subject: Fractions of an Amount
        Question: How do you say this fraction? $\frac{3}{5}$\\\\
        \prompt Answer 1: 3/5\\
        \prompt Answer 2: Three fives\\
        \prompt Judgment: Same\\\\
        \prompt Math Problem:```Subject: Multiplying and Dividing with Decimals\\
        \prompt Question: \( 50.09 \div 0.1= \)'''\\\\
        \prompt Answer 1: \( 500.09 \)\\
        \prompt Answer 2: 500.9\\
        \prompt Judgment: Different\\\\
    \prompt Math Problem:```Subject: Properties of Polygons\\
    \prompt Question: Tom and Katie are arguing about parallelograms. Tom says this shape is a parallelogram ![A four sided shape. All sides are equal, opposite angles are equal. There are no right angles.]() Katie says this shape is a parallelogram ![A four sided shape. Opposite sides are equal, all angles are right angles.]() Who is correct?'''\\
\prompt Answer 1: Katie\\
\prompt Answer 2: Only Katie\\\\
\prompt Judgment: Same\\
\prompt Math Problem:```Subject: Percentages of an Amount
Question: What is \( 120 \% \) of \( 50 \) ?'''\\\\
   \prompt  Answer 1: \( 10 \)\\
\prompt Answer 2: 60\\   
\prompt Judgment: Different\\\\
\prompt Math Problem:```Subject: Converting between Fractions and Percentages\\
\prompt Question: Convert this fraction to a percentage $\frac{4}{5}$'''\\\\
\prompt Answer 1: \( 45 \% \)\\
\prompt Answer 2: 45\\
\prompt Judgment: Same\\
        \bottomrule
    \end{tabular}
    \caption{Prompt to the \gptjudge model for judging answer equivalence for the distractor generation evaluations.}
    \label{tab:gpt_student_judge}
\end{table*}

\begin{table*}[t!]
    \renewcommand{\arraystretch}{0.8}
    \begin{tabular}{p{0.08\textwidth}p{0.86\textwidth}}
        \toprule
        \ttfamily \footnotesize 
        \textit{System} & \ttfamily \footnotesize You are an AI assistant designed to infer incorrect answers that students might give in solving K-12 math problems so that they can be used as distracter choices for multiple choice problems. You will be shown a math problem. Please output 3 incorrect answers. Your output should start with 'Incorrect Student Answers:' followed by a list of 3 incorrect answers that a student might give in solving the problem and that could be used as distracter choices for the question. Each item should have the same formatting as the correct answer, but make sure none of the incorrect answers you output are the exact same as the correct answer. Make sure to output exactly 3 incorrect answers. Do not output any other information. \\\\
        & \ttfamily \footnotesize Here are several examples of math problems and misconceptions for student answers. Please use these examples to guide your future responses. \\\\
        & \ttfamily \footnotesize Math Problem:```Subject: Adding and Subtracting Negative Numbers\\
        & \prompt Question: \( 6-(-8)= \)'''\\
        & \ttfamily \footnotesize Correct Answer: \( 14 \)\\
        & \ttfamily \footnotesize Incorrect Student Answers: \\
        & \prompt - \( -14 \) \\
        & \prompt - \( 2 \) \\
        & \prompt - \( -2 \) \\\\
        & \ttfamily \footnotesize Math Problem:```Subject: Multiplying and Dividing Negative Numbers\\
        & \prompt Question: \( -4+(-5) \times 3= \)'''\\
        & \prompt Correct Answer: \( 27 \)\\
        & \prompt Incorrect Student Answers:\\
        & \prompt - \( 27 \) \\
        & \prompt - \( -27 \) \\
        & \prompt - \( -11 \) \\\\
        & $\cdots$ \\\\
        & \ttfamily \footnotesize Math Problem:```Subject: Fractions of an Amount\\
        & \prompt Question: Here is a number card
        \( \frac{1}{3} \) of \( \mathrm{A} \) is \( 12 \) What is \( \frac{1}{2} \) of \( A \) ?\\
        & \prompt Correct Answer: \( 2 \)\\\\
        & \prompt Incorrect Student Answers: \\
        & \prompt - \( 2 \) \\
        & \prompt - \( 8 \) \\
        & \prompt - \( 36 \) \\\\
        & \ttfamily \footnotesize Math Problem:```Subject: Expanding Single Brackets\\
        & \prompt Question: The area of these shapes are equal. Which of these is correct? ![An image of two rectangles. The top one is blue and has the dimensions - side length: 4 and base 3x -2. The bottom one is yellow and has the dimensions - side length: 2 and base 5x + 2.]()'''\\
        & \prompt Correct Answer: \( 12 x-8=10 x+4 \)\\\\
        & \prompt Incorrect Student Answers: \\
        & \prompt - \( 12 x-2=10 x+2 \) \\
        & \prompt - \( 8+6 x-4=4+10 x+4 \) \\
        & \prompt - \( 4+3 x-2=2+5 x+2 \) \\\\
        \prompt User & \prompt Math Problem:```Subject: \textbf{Adding and Subtracting Fractions}\\
        & \prompt Question: \textbf{Work out: \( \mathbf{\frac{4}{11}+\frac{7}{11}} \). Write your answer in its simplest form.}'''\\
        & \prompt Correct Answer: \( \mathbf{1} \)\\\\
        \ttfamily \footnotesize {Assistant} & \prompt \textbf{Incorrect Student Answers: } \\
        & {- $ \mathbf{\frac{11}{22}}$ } \\
        & {- $\mathbf{\frac{11}{11}}$ } \\
        & {- $\mathbf{\frac{4}{11}-\frac{7}{11}}$ } \\
    \bottomrule
    \end{tabular}
    \caption{Example prompt/output for sampling incorrect answers with \unconditionalanswer. Bolded indicates variables that change.}
    \label{tab:unconditional_answer_prompt}
\end{table*}

\begin{table*}[t!]
    \renewcommand{\arraystretch}{0.8}
    \begin{tabular}{p{0.95\textwidth}}
        \toprule
        \prompt You are an AI assistant designed to simulate student misconceptions in solving K-12 math problems. You will be shown a problem and a misconception a student has in solving the problem. You should simulate what answer a student with that misconception would give. Your output should be the answer that the student would give. Please do NOT generate the correct answer. Instead, solve the problem as if you had the misconception that the student has.\\\\
        \prompt Please start your output by thinking step-by-step to reason through what answer a student with the misconception that you were shown would give. The step-by-step reasoning should start with "Reasoning:". \\\\
        \prompt Then, on a new line, output 'Incorrect Student Answer:' and generate the answer that the student would give. When outputting 'Incorrect Student Answer', please only output the letter of the multiple choice answer that the student would give and nothing else. Even if none of the multiple choice answers seem to be the right choice, output the letter of the closest answer. \\\\
        \prompt Here are several examples of math problems, misconceptions, and reasoning paths for student answers. Please use these examples to guide your future responses.\\\\
        \prompt Math Problem:```Subject: Range and Interquartile Range from a List of Data\\
        \prompt Question: What is the range of the following numbers? [0.2, \quad 0.22, \quad 0.212]\\
        \prompt Answer Choices:\\
        \prompt A) \( 0.02 \)\\
        \prompt B) \( 0.012 \)\\
        \prompt C) \( 0.21 \)\\
        \prompt D) \( 0.2 \)'''\\
        \prompt Student Misconception: When calculating the range does not reorder the data to find the largest number minus the smallest number\\\\
        \prompt Reasoning: Based on the student's misconception, we will assume they will not reorder the data and instead treat the first and last numbers in the list as the smallest and largest numbers. Then in order to calculate the range, they will subtract the smallest number from the largest number, i.e. 0.212-0.2=0.012. Thus, the student will choose answer choice B.\\
        \prompt Incorrect Student Answer: B\\\\
        \prompt Math Problem:```Subject: Gradient as change in y over change in x\\
        \prompt Question: What is the gradient of this line? ![A set of axes with a graph drawn on. The graph passes through the following points: (-3,-3) (-2,-2) (-1,-1) (0,0) (1,1) (2,2) and (3,3).]()\\   
        \prompt Answer Choices:
        \prompt A) \( 0 \)\\
        \prompt B) \( 2 \)\\
        \prompt C) \( x \)\\
        \prompt D) \( 1 \)'''\\
        \prompt Student Misconception: Believes the gradient of a line is given by an intercept\\\\
        \prompt Reasoning: If the student thinks the gradient of the line is given by the intercept, they will think 0 is the gradient because the intercept is 0. Therefore, they will choose A.\\
        \prompt Incorrect Student Answer: A\\\\
        $\cdots$
        \\\\
        \bottomrule
    \end{tabular}
    \caption{Prompt used for the \simulate module and student simulation evaluation experiments.}
    \label{tab:student_simulation_prompt}
\end{table*}

\begin{table*}[t!]
    \renewcommand{\arraystretch}{0.8}
    \begin{tabular}{p{0.95\textwidth}}
        \toprule
        \prompt You are an AI assistant designed to infer misconceptions that students might have in solving K-12 math problems. You will be shown a math problem, along with an incorrect answer given by a student. Please output a 1-sentence description of the misconception that a student who gave that incorrect answer might have. Make sure the misconception is specific enough that it could be used to infer the incorrect answer given by the student but general enough that it could apply to other similar problems.\\\\
        \prompt Your output should be formatted as follows:\\
\prompt Reasoning: [reasoning for the misconception]\\
\prompt Misconception: [1-sentence description of the misconception]\\
\prompt Do not output any other information.\\\\

\prompt Here are several examples of math problems and misconceptions for student answers. Please use these examples to guide your future responses.\\\\

\prompt Math Problem:```Subject: Range and Interquartile Range from a List of Data\\
\prompt Question: What is the range of the following numbers? [0.2, \quad 0.22, \quad 0.212]\\\\
\prompt Answer Choices:\\
\prompt A) \( 0.02 \)\\
\prompt B) \( 0.012 \)\\
\prompt C) \( 0.21 \)\\
\prompt D) \( 0.2 \)'''\\
\prompt Incorrect Student Answer: B\\\\

\prompt Reasoning: The student's answer, B, or 0.012, corresponds to the difference 0.212 - 0.2. This is the difference between the first and last numbers. Therefore, the student's misconception appears to be that they are not reordering the data to calculate the range.\\\\

\prompt Misconception: When calculating the range does not reorder the data to find the largest number minus the smallest number\\\\

\prompt Math Problem:```Subject: Gradient as change in y over change in x\\
\prompt Question: What is the gradient of this line? ![A set of axes with a graph drawn on. The graph passes through the following points: (-3,-3) (-2,-2) (-1,-1) (0,0) (1,1) (2,2) and (3,3).]()\\

\prompt Answer Choices:\\
\prompt A) \( 0 \)\\
\prompt B) \( 2 \)\\
\prompt C) \( x \)\\
\prompt D) \( 1 \)'''\\

\prompt Incorrect Student Answer: A\\\\

\prompt Reasoning: The student answers A, or 0, and the x and y intercepts of the given line are also both 0. Therefore, the student's misconception may be about confusing the gradient of the line with one of the intercepts.\\\\

\prompt Misconception: Believes the gradient of a line is given by an intercept\\\\

\prompt Math Problem:```Subject: Length Units\\
\prompt Question: 450 cm = $\square$ m\\\\

\prompt Answer Choices:\\
\prompt A) \( 4.5 \)\\
\prompt B) \( 45 \)\\
\prompt C) \( 5 \)\\
\prompt D) \( 45000 \)'''\\

\prompt Incorrect Student Answer: C\\\\

\prompt Reasoning: To find the value of the square, the correct approach would be to convert $[450 cm]$ to meters. There are 100 centimeters in a meter, and so the correct calculation is $[450 \div 100 = 4.5 m]$. However, the student answers C, or 5, which suggests that they may be rounding up to the nearest whole number to give their answer.\\\\

\prompt Misconception: Gives a rounded whole number instead of a decimal\\
$\cdots$
\\
        \bottomrule
    \end{tabular}
    \caption{Prompt used for the \infer module and for the misconception inference experiments.}
    \label{tab:misconception_inference_prompt}
\end{table*}

\begin{table}[h]
\centering
\begin{tabular}{lcc}
\toprule
\multirow{3}{*}{\textbf{Variant}} & \multicolumn{2}{c}{\textbf{Values of $\alpha$ when examples are:}}\\
& \textbf{cycle consistent} & \textbf{correct answer} \\
& ($\answertwo = a$) & ($s = \correctanswer$) \\
\midrule
\methodfullvariant & \textbf{2} & \textbf{0} \\
\methodcorrectvariant & 1 & \textbf{0} \\
\methodnocycle & 1 & 1 \\
\bottomrule
\end{tabular}
\caption{Variants of \method and the values of $\alpha$ that they use, where $\alpha$ the weight given to each example. $\alpha=1$ is the default value.}
\label{tab:variants}
\end{table}

\begin{table*}[t!]
    \renewcommand{\arraystretch}{0.8}
    \begin{tabular}{p{0.95\textwidth}}
        \toprule
        \prompt Model outputs for \textbf{student simulation} (validation dataset, round 4).\\\\
        
        \prompt Math Problem:```Subject: Volume and Capacity Units\\
        \prompt Question: 3 litres = $\color{gold}\star$ millilitres\\
        \prompt Answer Choices: A) 0.3 \quad B) 30 \quad C) 300 \quad D) 3000'''\\\\
        \prompt Student Misconception: Thinks there are 100 ml in a litre\\\\
        \prompt Model Output: Reasoning: Assumes 100 ml per litre, so $3 \times 100 = 300$. Chooses C.\\\\
        \prompt Incorrect Student Answer: C\\\\
        \prompt Actual Output: C\\\\
        \midrule
        
        \prompt Math Problem:```Subject: Multiples and Lowest Common Multiple\\
        \prompt Question: Which of the following numbers is a common multiple of $4,6,$ and $12$?\\
        \prompt Answer Choices: A) 2 \quad B) 12 \quad C) 18 \quad D) 20'''\\\\
        \prompt Student Misconception: Identifies a multiple of a number but not a common multiple of a group\\\\
        \prompt Model Output: Reasoning: Thinks 12 is a multiple of 4, 6, and 12. Chooses B.\\\\
        \prompt Incorrect Student Answer: B\\\\
        \prompt Actual Output: D\\\\
        
        \bottomrule
    \end{tabular}
    \caption{Example model outputs for student simulation from the validation dataset in round 4.}
    \label{tab:examples_student_sim}
\end{table*}

\begin{table*}[t!]
    \renewcommand{\arraystretch}{0.8}
    \begin{tabular}{p{0.95\textwidth}}
        \toprule
        \prompt Model outputs for \textbf{misconception inference} (validation dataset, round 4).\\\\
        
        \prompt Math Problem:```Subject: Expanding Double Brackets\\
        \prompt Question: What should replace the circle when these two brackets are expanded and simplified?\\
        $(p-3)(p+1) \equiv p^{2} \triangle p \bigcirc$\\
        \prompt Answer Choices: A) $\bigcirc=-3$ \quad B) $\bigcirc=+3$ \quad C) $\bigcirc=-4$ \quad D) $\bigcirc=-2$'''\\\\
        \prompt Incorrect Student Answer: B ($+3$)\\\\
        \prompt Model Inferred Misconception: Neglects to distribute the negative sign when expanding brackets\\\\
        \prompt Actual Misconception: Believes multiplying a positive by a negative gives a positive answer\\\\
        \midrule
        
        \prompt Math Problem:```Subject: Adding and Subtracting with Decimals\\
        \prompt Question: $50.09 - 0.1 =$\\
        \prompt Answer Choices: A) 49.99 \quad B) 50.99 \quad C) 50.08 \quad D) 38.98'''\\\\
        \prompt Incorrect Student Answer: B ($50.99$)\\\\
        \prompt Model Inferred Misconception: Thinks subtraction is performed as addition\\\\
        \prompt Actual Misconception: When borrowing in subtraction, does not subtract one from the preceding digit\\\\
        
        \bottomrule
    \end{tabular}
    \caption{Example model outputs for misconception inference from the validation dataset in round 4.}
    \label{tab:examples_misconception_inference}
\end{table*}

\end{document}